\documentclass[letterpaper, 10 pt, conference, compress]{ieeeconf}  
\usepackage{graphicx} 
\usepackage{xcolor, soul}
\usepackage{cite}
\usepackage{amsmath}
\usepackage{siunitx}
\usepackage{amssymb}
\usepackage{float}
\usepackage{comment}

\IEEEoverridecommandlockouts                              

\title{\LARGE \bf
Imitation Learning for Adaptive Control of a Virtual Soft Exoglove}

\author{Shirui Lyu$^{1}$, Vittorio Caggiano$^{2}$, Matteo Leonetti$^{1}$, Dario Farina$^{3}$, and Letizia Gionfrida$^{1, 4} $
\thanks{*This work was supported by The Royal Academy of Engineering Research Fellowship (RF2324-23-229).}
\thanks{$^{1}$S. Lyu, M.Leonetti and L. Gionfrida are with the Department of Informatics, King's College London, London, WC2R 2LS, UK.
        {\tt\small {(shirui.lyu, matteo.leonetti, letizia.gionfrida)}@kcl.ac.uk}}%
 \thanks{$^{2}$V. Caggiano is with MyoLab, New York, USA and Harvard Medical School and Spaulding Rehabilitation Hospital, Boston, USA
         {\tt\small caggiano@gmail.com}}%
\thanks{$^{3}$D. Farina is with the Department of Bioengineering, Faculty of Engineering, Imperial College London, London, SW7 2AZ, UK
{\tt\small d.farina@imperial.ac.uk}}%
\thanks{$^{4}$L. Gionfrida is also with John A. Paulson School of Engineering and Applied Sciences, Harvard University, Cambridge, MA, USA.
{\tt\small 
}}%
}

\begin{document}

\maketitle
\thispagestyle{empty}
\pagestyle{empty}

\begin{abstract}
The use of wearable robots has been widely adopted in rehabilitation training for patients with hand motor impairments. However, the uniqueness of patients' muscle loss is often overlooked. Leveraging reinforcement learning and a biologically accurate musculoskeletal model in simulation, we propose a customized wearable robotic controller that is able to address specific muscle deficits and to provide compensation for hand-object manipulation tasks. Video data of a same subject performing human grasping tasks is used to train a manipulation model through learning from demonstration. This manipulation model is subsequently fine-tuned to perform object-specific interaction tasks. The muscle forces in the musculoskeletal manipulation model are then weakened to simulate neurological motor impairments, which are later compensated by the actuation of a virtual wearable robotics glove. Results shows that integrating the virtual wearable robotic glove provides shared assistance to support the hand manipulator with weakened muscle forces. The learned exoglove controller achieved an average of 90.5\% of the original manipulation proficiency.


\end{abstract}

\section{INTRODUCTION}

The use of wearable robots has shown promising potential both in neurological rehabilitation protocols \cite{hatem2016rehabilitation} and in passive supports to boost neural motor performance \cite{nuckols2021individualization}. Patients with upper limb neurological dysfunctions or impairments often experience varying degrees of dexterity loss \cite{gassert2018rehabilitation}. A personalized rehabilitation strategy is essential to enhance their independence, thereby improving their quality of life \cite{rodgers2017stroke}. However, designing a tailored rehabilitative intervention that accounts for rapidly changing, patient-specific clinical indicators \cite{richards2015stroke}, particularly through wearable robotic devices, remains a significant challenge. 

One of the challenges in realizing an intelligent controller for assistive wearable robots is the translation of the unique medical conditions of individual patient into adaptable controllers. For example, patients recovering from ischemic or hemorrhagic strokes, often experience significant limitations in their range of motion, characterized by symptoms such as reduced muscle strength or limitation of range of motion\cite{azzollini2021does}. 
In clinical settings, such diverse medical conditions are typically managed through ad-hoc manual rehabilitation programs administered under the supervision of trained professionals\cite{rodgers2017stroke}. However, in the context of wearable robotic rehabilitation, the device controller needs to be timely adjusted to provide clinical support \cite{kwakkel2003probability, nishimura2007time}. The development of automated methods for adapting wearable device control to individual patient needs is crucial for enabling personalized and accessible rehabilitation interventions.

Data-driven control approaches for wearable robotics have become widely adopted, enabling more general and accessible solutions for wearable robotic control\cite{lotti2020adaptive, zhang2023real, li2024data, simonetti2024wearable, molinaro2024task, tricomi2024soft}. These approaches learn biological features of the user such as kinematics, and muscle dynamics, from sufficient data to control the assistive devices. However, they focus on general rather than specific user behaviours, thus leaving a gap in addressing practical individual-based rehabilitation needs.

The development of personalized control strategies for wearable robots has been employing musculoskeletal models to facilitate adaptable controller training through Reinforcement Learning (RL) \cite{hodossy2023shared, luo2024experiment}. Despite their utility, these approaches often yield non-intuitive behaviours \cite{chen2024vividex} attributable to the absence of reliable human-reference data. To address these limitations, the integration of vision and imitation learning within RL training frameworks offers a promising solution. These methods enhance the fidelity of natural human behaviour modelling by providing enhanced biomechanical insights.

In this work, we propose a framework that uses imitation learning to derive user-specific reference trajectories from a healthy subject’s video dataset  \cite{dexycb-chao-2021}. These trajectories are used to reconstruct kinematics and muscle dynamics in silico, where hand impairments are simulated. Finally, a tendon-driven exoskeleton glove (exoglove) control strategy that adapts to the simulated impairments is tested using the healthy reference as a basis for optimization reference.

\begin{figure*}[ht]
    \centering
    \includegraphics[width=1\textwidth]{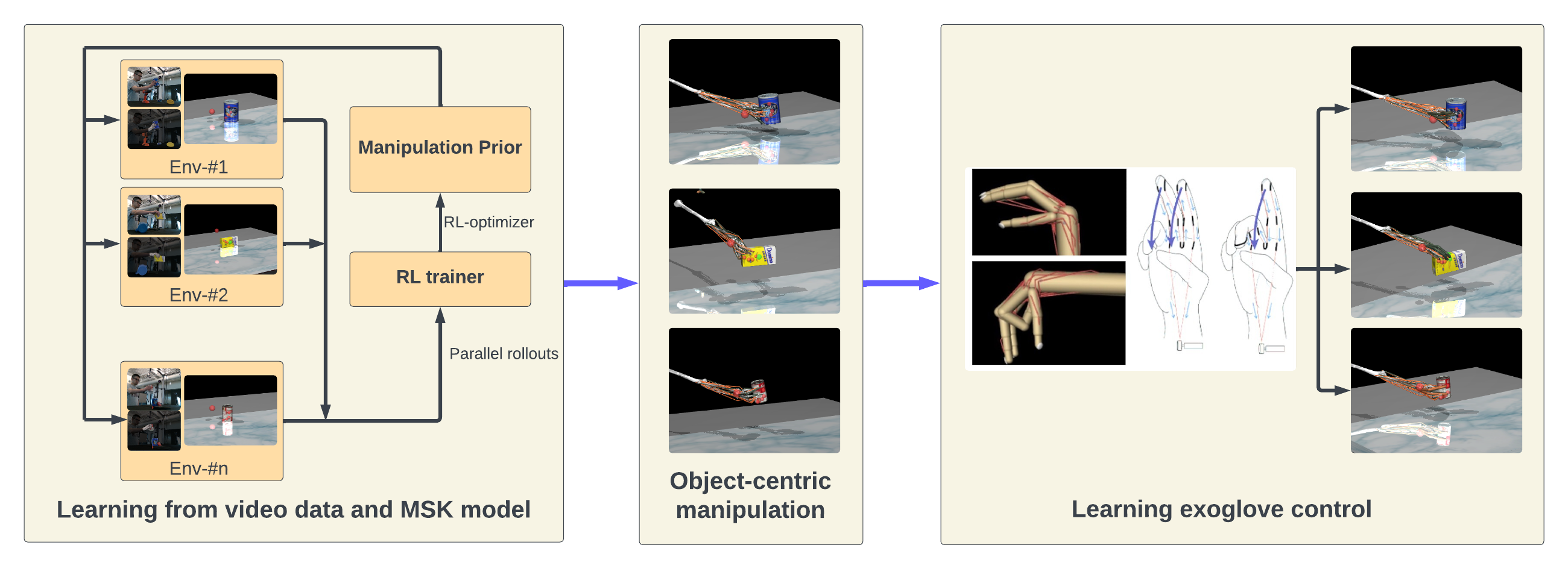}
    \caption{Pipeline overview illustrating the three main stages: (1) extraction of user behavior via imitation learning from healthy subject video data, (2) training a biologically accurate musculoskeletal hand model for object-centric manipulation and simulating muscle impairments, and (3) development and testing of a tendon-driven exo-glove controller to compensate for impaired hand functionality.}
    \label{fig:main-pipeline}
\end{figure*}

The proposed approach consists of three stages (Fig.\ref{fig:main-pipeline}). First, reference trajectories are extracted from video data and learned by imitating the behaviour of a healthy subject through \textit{MyoHand} \cite{myosuite-caggiano-2022}, a biologically accurate musculoskeletal hand model, to replicate the user's natural behaviour. Second, RL-driven manipulation is applied to simulate hand-object manipulation tasks within the musculoskeletal model, and then a simulated impairment is incorporated to reflect reduced functionality in the hand. Finally, a virtual exoglove, modeled after a real device \cite{rho2021learning}, is implemented in simulation, where an adaptive control strategy is developed to compensate for the impaired hand, optimizing hand-object manipulation and improving functionality for the impaired hand.

\section{Related Work}

\ifpdf
    \graphicspath{{Chapter2/Figs/Raster/}{Chapter2/Figs/PDF/}{Chapter2/Figs/}}
\else
    \graphicspath{{Chapter2/Figs/Vector/}{Chapter2/Figs/}}
\fi

\subsection{Learning manipulations from human videos and RL}
As human video inherently captures the intricate kinematics of hand-object interactions, including the complexity and precision required for dexterous manipulation, extensive research has focused on understanding and modeling these interactions \cite{GRAB:2020, Liu_2022_CVPR, dexycb-chao-2021, hasson2019learning}. Those recorded interactions provide sufficient context for the understanding of dexterous manipulation and facilitate other robotics applications such as handling previously unseen manipulation tasks \cite{chane2023learning}. Nevertheless, the learned model, referring to computational models trained on human video data to predict or replicate manipulation strategies, depends heavily on human video data and is less robust in handling unseen objects.

RL-based manipulation, on the other hand, is often data-inefficient and explores solutions with computational resources over data. Such an approach gives rise to manipulation understanding that is less relevant to human demonstrations. Object-centric approaches \cite{pgdmdasari-2023, caggiano2023myodex}, for example, mainly match the object in hand with a reference trajectory, and learn a manipulation strategy even with a high Degree of Freedom (DoF) musculoskeletal hand model.

Recent work \cite{qin2022dexmv, mandikal2022dexvip, shaw2023videodex} has attempted to combine the advantage from both sides: by using human demonstration video to facilitate RL training, and by using RL to train robust grasping behaviours. In addition, Wan et al. \cite{wan2023unidexgrasp++} leveraged RL and computer vision to first generate multiple grasping gestures before applying conditioned policy training to achieve generalized grasping behaviour. Chen et al. \cite{chen2024vividex} proposed to use of human demonstration video to guide the RL training while leaving space for RL exploration to handle robot-object interactions. However, most previous works generalize the grasping behaviour and few of the studies were conducted on model-based RL for rehabilitative robotics applications.


\subsection{Model-based wearable robotics control}

Recent studies\cite{tricomi2024soft, zhang2023real, li2022human, simonetti2024wearable}, on wearable robotics systems, have focused on integrating high-level task control with low-level hardware control to enhance system performance. This combined approach allows wearable robots to execute tasks more effectively while adapting to the mechanical and dynamic constraints of the hardware. Introducing high-level perceptual information such as egocentric views \cite{gionfrida2024wearable, kim2019eyes}, or user-intent \cite{hodossy2023shared}, into the control architecture usually improves the system's responsiveness and adaptability. 
However, deriving accurate and reliable perceptual information requires meticulous modelling of biological states. These models must be biologically accurate to ensure that the system's responses align appropriately with the user's physiological and cognitive conditions, thereby enhancing the overall efficacy and safety of the wearable robot. Hodossy and Farina\cite{hodossy2023shared} proposed to use a humanoid model and motion tracking to learn a robust locomotion model to inform the training of an intention-aware prosthesis. Luo et al. \cite{luo2024experiment} leveraged a human musculoskeletal model to train an exoskeleton to provide locomotion support based on the user states. However, few studies have focused on upper-limb wearable robotics aimed at providing shared assistance through musculoskeletal hand models.


\section{Methodology}
Our approach consists of three main stages. First, we learn the user's behavior context by extracting joint kinematics from a video dataset \cite{dexycb-chao-2021} of a healthy subject. Second, a biologically accurate musculoskeletal hand model is trained for object-centric manipulation tasks, enabling the simulation of hand-object interactions that reflect the user's behaviour. In this stage, the model is also adapted to simulate a designated muscle-weakened hand impairments. Lastly, a tendon-driven soft exoglove is applied to the musculoskeletal hand, with an RL-based glove controller, to support dexterous manipulation of the impaired hand model. The trained controller is then tested on a muscle-weakened hand model to evaluate its effectiveness in assisting with grasping actions, compensating for the impaired hand’s functionality. The pipeline overview is illustrated in Fig.\ref{fig:main-pipeline} .

\subsection{Learning behaviour context from video data}

Leveraging video data of a subject grasping daily objects \cite{dexycb-chao-2021}, we first learn the subject's behaviour context by distilling them into a musculoskeletal hand model controller via RL-based imitation learning. Three objects of varying size, including a master-chef can, a tomato can and a sugar box from the YCB\cite{calli2015ycb} dataset are used to evaluate the learned hand-object manipulation. The reference grasping type on those objects resembles two commonly used grasp taxonomy types \cite{feix2015grasp}, power-grasp and precision-grasp, with variations in object sizes. The hand model observes the reference joint positions and learns a mapping to control the hand's muscle activation signals to follow the reference trajectory. Similar to a previous study \cite{caggiano2023myodex}, a prior model is first trained to capture the essence of the subject's general behaviour context by extracting a joint experience from a number of reference trajectories but from the same subject for training efficiency. The behaviour prior is later fine-tuned to individual hand-object interaction tasks for accuracy. We have observed that learning such behaviour prior can contribute to faster training convergence on previously unseen trajectories and objects.
The reward function of such prior model training only comprises the degree of matching of the reference trajectory with the rough gesture shape, similar to the approach by Chen et al. \cite{chen2024vividex}. Nevertheless, we increased the weighting of the thumb tip, to balance it with the other 4 fingertips to follow the reference gesture.

\subsection{Musculoskeletal object manipulator}
Imitation learning from video references effectively models user behaviours by capturing hand joint kinematics. However, it does not account for critical hand-object interaction details, such as the contact information for stable grasping and manipulation. To model the contact between hand and object, an RL-based manipulator is employed on the Mujoco \cite{Todorov-mujoco-2012} physics simulator together with a existing musculoskeletal hand model from Myosuite \cite{myosuite-caggiano-2022}.

In addition, clinical tests such as the Box and Block Test \cite{mathiowetz1985adult} and the Action Research Arm Test \cite{yozbatiran2008standardized} are common evaluations of dexterous proficiency of patients with upper-limb motor impairment. Such tests involve moving objects along a required trajectory as in most routine tasks. Thus, we define our exoglove controller training via an object-centric \cite{pgdmdasari-2023} RL approach, namely manipulating an object following a predefined trajectory to align with clinical rehabilitative requirements.

\textbf{Problem definition}

Similar to a previous work \cite{pgdmdasari-2023} on RL-based object-centric manipulation, the problem is defined as a Markov Decision Process (MDP) and solved via reinforcement learning. The Markov decision process is defined as a tuple $  M = (S, A, T, R, \rho, \gamma)$, where $ S \subseteq \mathbb{R}^n$ and $ A \subseteq \mathbb{R}^m $ represent the continuous state space of observations and the action space respectively. The unknown transition dynamics are described by \( s' \sim T(\cdot|s, a) \). In addition, \( R : S \rightarrow [0, R_{\text{max}}] \), denotes the reward function, \( \gamma \in [0, 1) \) denotes the discount factor, and \( \rho \) the initial state distribution. In RL, a policy is a mapping from states to a probability distribution over actions, i.e. \( \pi : S \rightarrow P(A) \), which is parameterized by \( \theta \). The goal of the agent is to learn a policy \( \pi_\theta(a|s) = \arg\max_\theta[J(\pi,M)] \), where \( J = \max_\theta \mathbb{E}_{s_0 \sim \rho(s),a \sim \pi_\theta(a_t|s_t)} \left[\sum_t R(s_t, a_t)\right] \).

\textbf{State space}

The state space for the musculoskeletal hand model and the exoglove consists of $\{\phi, \psi, \tau, \hat{\theta} \}$. $\phi$ a 58-dimensional vector of 23 hand and 6 arm joints positions and velocity of the musculoskeletal model. $\psi$ is the object pose (3-dimensional), orientation (3-dimensional) and velocity (3-dimensional). $\tau$ is a positional encoding used to represent time steps in the tasks to handle possible repetitions \cite{vaswani2017attention}. In addition, the reference motion $\hat{\theta}$ is incorporated as a 63-dimensional vector to indicate the joint position of the wrist, and the four finger joints, including the metacarpophalangeal (MCP), the distal interphalangeal (DIP), proximal interphalangeal (PIP) joints and the fingertip from the human demonstrations. 

\textbf{Action space}
 
The action space for the manipulation baseline is a 45-dimensional vector that consists of continuous activations for 39 muscles of the upper right arm, together with 3D translation and rotation of the shoulder, as in MyoDex \cite{caggiano2023myodex}.

The action space for the exoglove controller is a 3-dimensional vector representing finger contraction and extensions applied on the index and middle finger, with one extra dimension for thumb support.

\textbf{Reward function}

Similar to a previous work by Chen et al. \cite{chen2024vividex}, we define two reward functions to support the training. The first function is defined on the object center of mass matching with the one from the reference trajectory, and a second function is defined as matching the musculoskeletal hand joint positions with those from the human demonstration. 

For object position $x^{(p)}_t$ and orientation $x^{(o)}_t$ at time step $t$, the reward function for matching the object position and orientation to the reference trajectory is defined by:

$$
\textbf{R}_{obj} := \lambda_1 exp\bigl\{-\alpha_1 ||x^{(p)}_t - \hat{x}^{(p)}_t||_2 - \beta|\angle x^{(o)}_t - \angle \hat{x}^{(o)}_t| \bigl\}
$$

The reward function for matching the joints' position with those from the human demonstration is:

$$
\textbf{R}_{demo} := -\lambda_2 \bigl\{\alpha_2 ||x^{(p)}_t - q(t)||_2 \bigl\}
$$

where $(\hat{x}^{(p)}_t, \hat{x}^{(o)}_t)$ are the object's reference trajectory's position and orientation at time $t$, and $q(t)$ denotes the reference joints position from human demonstration data at time $t$. Here we selected the fingertips of all five fingers and the wrist to accommodate the musculoskeletal hand model. In addition, we increased the weighting of the thumb tip to balance the gesture with reference tracking accuracies. The constants $\alpha, \beta$ and $\lambda$s are part of training hyper-parameters, following the same set-up as in MyoDex \cite{caggiano2023myodex}.


In terms of RL training, both the behaviour prior model that tracks human demonstration and the object-oriented manipulator are equipped with the same state space and action space. Yet the reward function applied to train the two models is slightly different. The prior model is trained by collecting parallel roll-out from a number of manipulation tasks with human demonstrations, each of which contains a different human demonstration, and the reward used is purely $\textbf{R}_{demo}$ to maximize the following of trajectory. On the other hand, the manipulation model is trained on a reward function defined by $\textbf{R} = \textbf{R}_{demo} + \textbf{R}_{obj}$, to take account of the influence from both the video demonstration and object manipulation for each time step in the training environment and are subject to the specific manipulation tasks.

\textbf{Object centered interaction}

To achieve a realistic interaction modelling with the musculoskeletal model, the hand-object interaction is then trained by fine-tuning the prior model from human demonstrations on individual manipulation tasks. In each of the simulation environments, the controller is required to move the object to follow a predefined reference trajectory, extracted from the human demonstration dataset, and learns a successful manipulation strategy for each object and reference trajectory. To reduce the complexity of the tasks, the controller is trained on top of the human demonstration prior, which was capable of following the human references to facilitate training. In the interaction tasks, the musculoskeletal controller's objective is to find a balance between following the reference and the object manipulation proficiency.

\subsection{Development of the exoglove controller}

\begin{figure}[ht]
\centering
\includegraphics[width=0.48\textwidth]{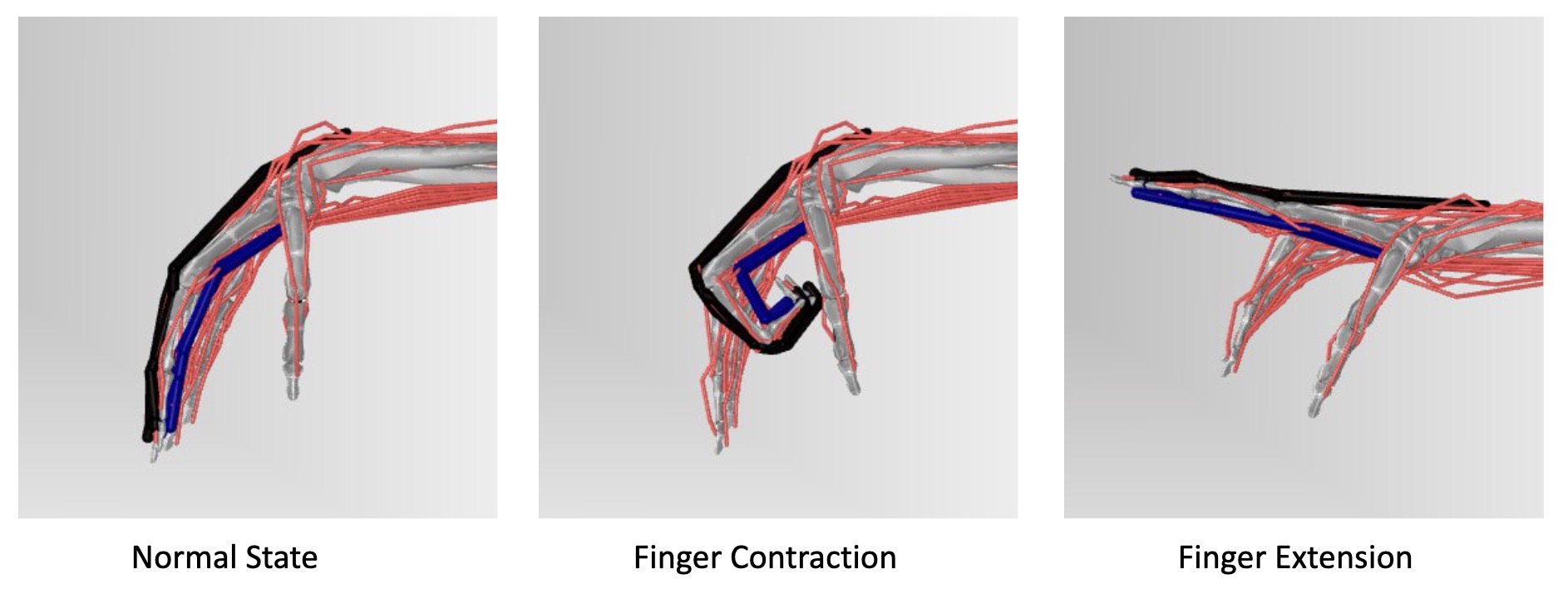}

\caption{Visualization of the tendon-driven exoglove actuation in simulation, based on a real prototype \cite{rho2021learning}. This figure visualizes the exoglove actuation mechanism, from left to right are the hand in normal state, finger contraction, and extension.}

\label{chap3:exoglove-sim}
\end{figure}

On top of the manipulation baseline, a tendon-driven exoglove based on a real prototype \cite{rho2021learning} was implemented in the simulation. Yet the simulated exoglove has an extended degree of freedom (DoF) with a total of 3 controlling dimensions: finger contraction, extension and passive thumb support. To align with the design of the physical prototype, the glove actuation is only applied to the index and middle fingers. The following (Fig. \ref{chap3:exoglove-sim}) shows the actuation mechanism of the glove in simulation. The thumb support dimension simulates passive support to the palm and only provides internal rotation.

\textbf{Learning compensatory movements}

Deriving a suitable, personalized exoglove controlling strategy based on the user behaviour context is challenging. To train such a shared assistive exoglove controller, we adopt the same object-centric approach to support the manipulation behaviour. The objective of such an approach is to match the interacted object coordinates and orientation with a reference trajectory in simulation, which aligns with the needs of motor-impaired users.

Similar to recent work on RL-based prosthesis control \cite{hodossy2023shared}, the glove controller is equipped with the same set of training hyper-parameters, reward and observation space as for the musculoskeletal hand-on-object manipulation. The aim of such a design is to align the glove controller and musculoskeletal hand model with the same objective of learning how to maximize shared assistance. This design also ensures the information observed by the glove controller is suitable for practical applications, as those features can be extracted via well-researched computer vision algorithms.






\section{Experiments}





\subsection{Learning behaviour context from video data}

The human behaviour reference data used in this experiment is derived from a video-based dataset DexYCB \cite{dexycb-chao-2021}. The dataset includes a subject manipulating different YCB objects \cite{calli2015ycb}, providing the hand joint positions as extracted via computer vision. In our case, the reference trajectories are defined by the extracted joint kinematics, and the data to learn such behaviour all come from the same subject (with ref code: 20200709-subject-01 \cite{dexycb-chao-2021}). Then the training environment for the musculoskeletal model is recreated from the reference dataset using the Myosuite \cite{myosuite-caggiano-2022} simulator for efficient contact-rich computations.

\begin{figure}[ht]
    \centering
    \includegraphics[width=0.49\linewidth]{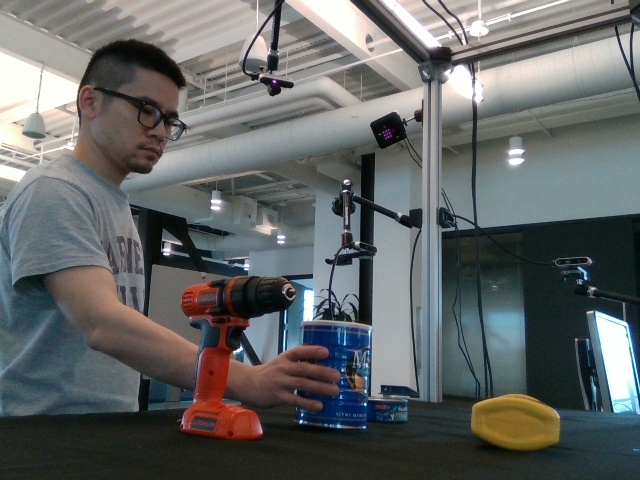}
    \includegraphics[width=0.49\linewidth]{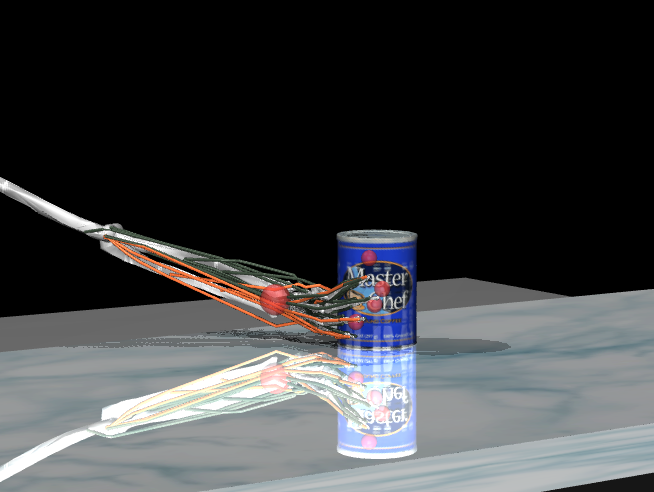} 
    \vspace{0.5cm}
    \includegraphics[width=0.49\linewidth]{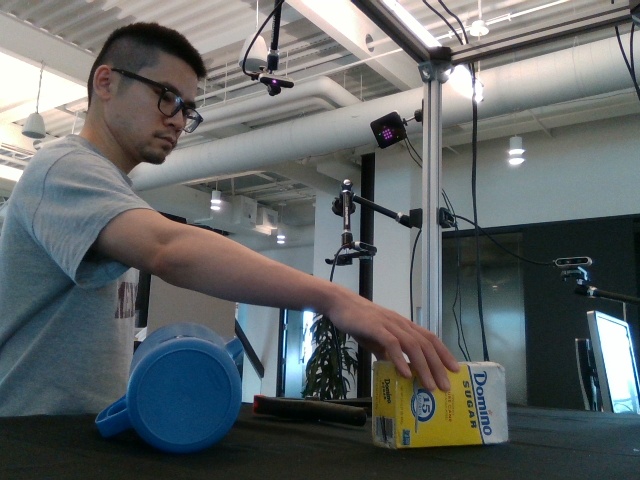}
   \includegraphics[width=0.49\linewidth]{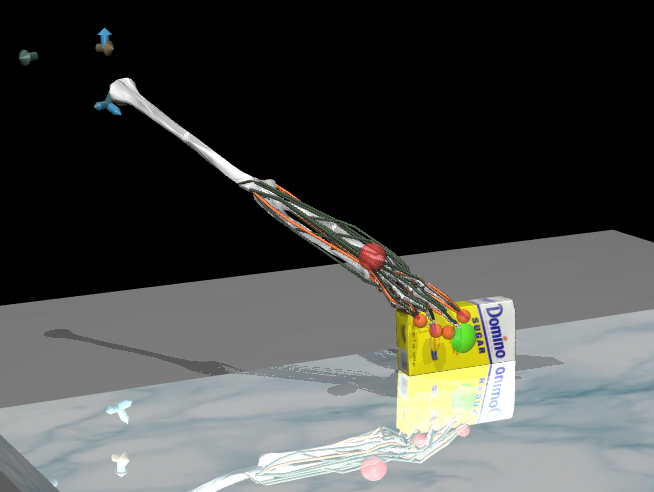}
    \caption{Visualization of the learned musculoskeletal model policy. The policy is initially trained as a motion prior, capturing general trajectory tracking capabilities, and is subsequently fine-tuned for specific downstream tasks. The figure illustrates the learned prior policy applied to manipulating a chef can (top row) and a sugar box (bottom row).}
    \label{fig:hdemo-prior}
\end{figure}
\begin{figure}[!ht]
    \centering
    \includegraphics[width=0.95\linewidth]{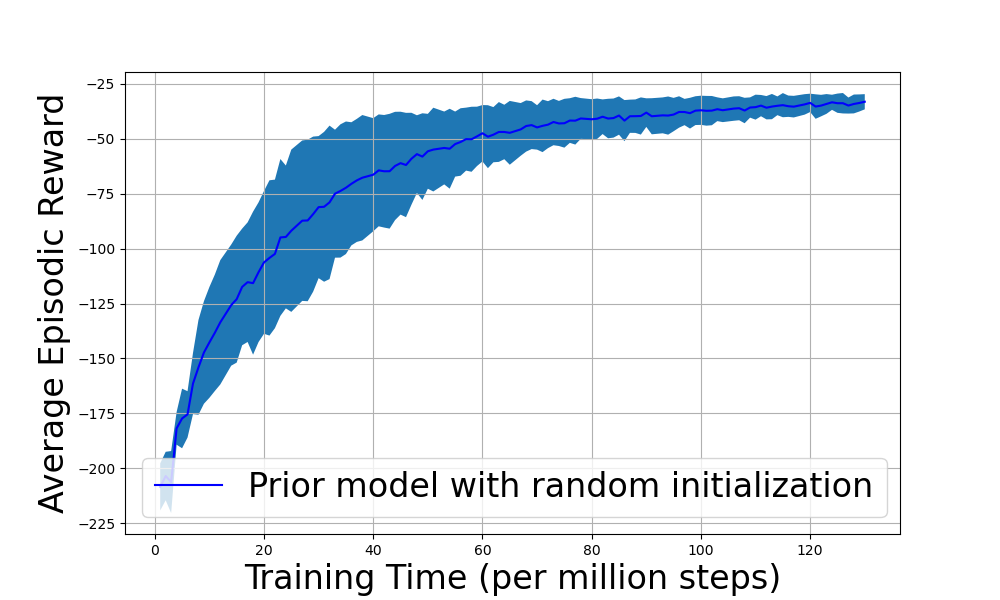}
    \caption{Prior model training with random initialization}
    \label{fig:prior-train}
\end{figure}

To extract the behavioral context from the video data, a control agent is initially trained as a motion prior to establish general trajectory tracking capabilities with the musculoskeletal hand model. This agent is then fine-tuned to accurately replicate the specific trajectories of the individual user. A total of 16 environments of extracted reference joint kinematics are incorporated in simulation, including the interaction with 6 objects: chef can, tomato can, tuna can, mustard bottle, pudding box and sugar box. The RL agent learns from the joint experience of the 16 environments by collecting their parallel rollout to control the musculoskeletal hand model. The policy is trained until convergence as a motion prior and is then fine-tuned on individual downstream tasks. Fig. \ref{fig:hdemo-prior} shows the learned prior policy interacting with a chef can and a sugar box. Regarding model robustness, a random Gaussian noise with zero mean and 3\% standard deviation is included in the observation space on hand joints and object kinematics in the prior training. We also test the model performance with random initialization (Fig. \ref{fig:prior-train}).

\begin{figure}[!ht]
    \centering
    \includegraphics[width=0.95\linewidth]{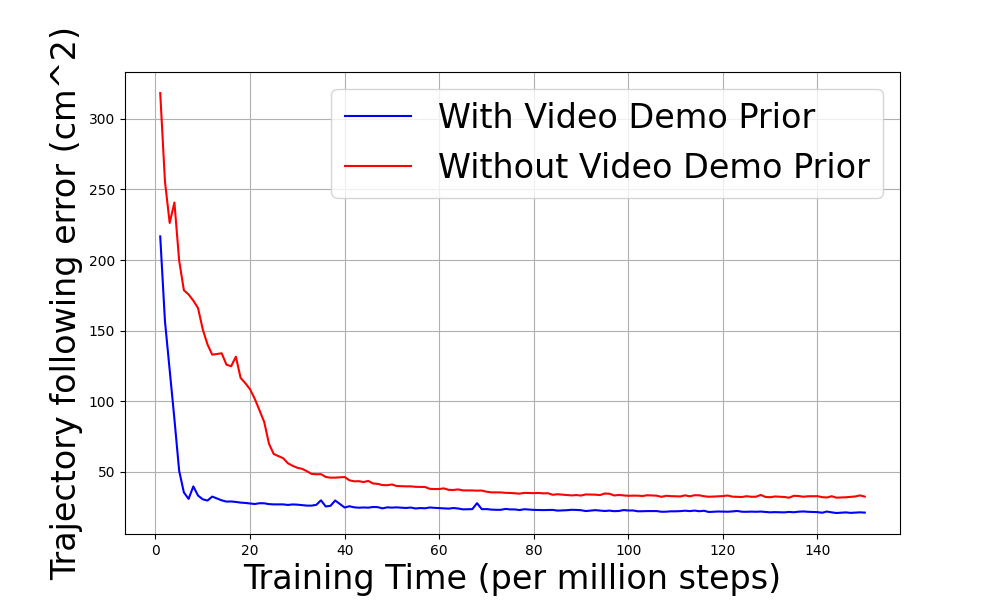}
    \caption{Comparison of model training \textbf{WITH} (blue) and \textbf{WITHOUT} (red) the learned behaviour prior on previously unseen object and trajectory pair.}
    \label{fig:curve}
\end{figure}

The learning of a prior model is a common approach for tasks with high DoF\cite{caggiano2023myodex}. In this case, we show that this prior model improves the time to convergence for previously unseen reference trajectories from the same subject's data (Fig. \ref{fig:curve}). The trajectory following error is defined as the absolute value of the average $\textbf{R}_{demo}$ reward over a training episode. The value serves as an indicator of the spatial difference between the position from human reference and the observed position of the musculoskeletal model. This shows that learning from a collection and diverse range of manipulation video data helps better understand the user's behaviour context.

\subsection{Training of hand-object manipulation tasks}
\label{sec:manipulation-task}

Leverage demonstration prior, a musculoskeletal hand controller is fine-tuned to learn the interaction with specific object and trajectory pairs. For each object-trajectory pair, a single control policy is learned on top of the prior model and the same set of hyper-parameters from MyoDex \cite{caggiano2023myodex} is given to the RL algorithm. However, the reward function used for object manipulation is the sum of $\textbf{R}_{demo}$ and $\textbf{R}_{obj}$. The manipulation model is then trained for 26 million environment steps until fully converge. For each step during evaluation, the object's centre of mass has to be within a 2.5-cm range from the reference position to be considered a successful step. The success rate is computed by dividing successful steps over the total steps. Previous literature\cite{pgdmdasari-2023} has noted that human perceives a task as successful if the success rate exceeds 80\%. The outcome of the policy training (Fig. 4) and the average success rate over 30 trials on individual objects are recorded in Tab. \ref{tab:succ-obj}.

\begin{table}[ht]
\centering
\caption{Success Rates under 26 million steps}\label{tab:succ-obj}
\begin{tabular}{ c c c c c}
\hline
\textbf{Objects} & \textbf{Chef can} &  \textbf{Sugar box}&  \textbf{Tomato can} \\ \hline
Performances       & 89\%       & 97.6\%  & 93.8\%
 \\ 
\hline \hline
\end{tabular}
\end{table}

\subsection{Learning of exoglove control}




\begin{figure}
    \centering
    \includegraphics[width=0.95\linewidth]{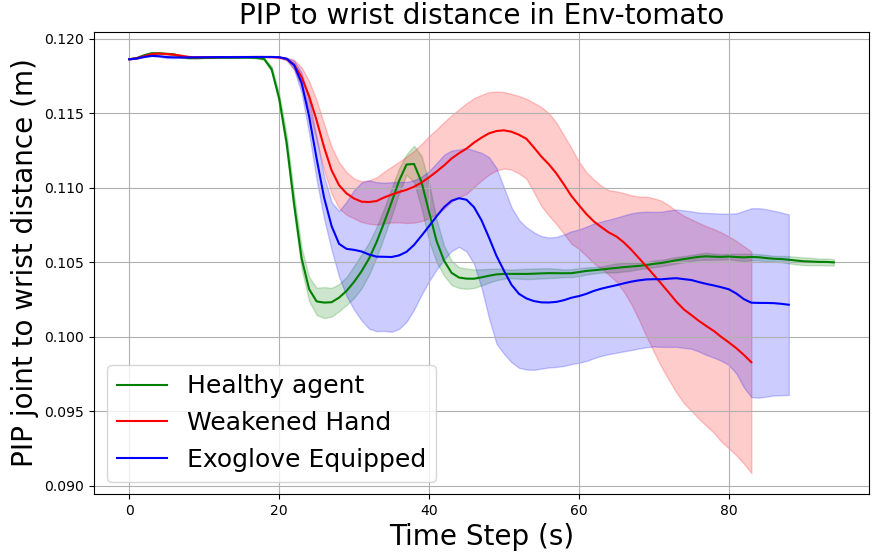}
    \caption{Comparison of hand actuation behaviours for the tomato can, illustrating the reduction in distance from the proximal interphalangeal (PIP) joint to the wrist during the contraction phase for the healthy baseline (green), weakened hand without assistance (red), and weakened hand with exoglove compensation (blue). The mean and the standardization of the middle finger PIP joint to the wrist are recorded from a total of 30 trials. The figure depicts the exoglove's functionality in compensating for the weakened hand's reduced performance
 .}
    \label{fig:glove-mech}
\end{figure}


To simulate neurological muscle weakness, the sarcopenia model from Myosuite \cite{myosuite-caggiano-2022} is introduced to reduce the maximum muscle strength of the musculoskeletal hand. Then, on top of the weakened hand model, the exoglove controller is trained to compensate for the impaired hand's movements via shared assistance. With the weakened musculoskeletal hand model frozen, the exoglove controller is trained to learn a compensatory movement based on the hand joint kinematics to provide shared assistance. The exoglove controller is equipped with the same observation space and reward function as the musculoskeletal hand, an approach to ensure alignment in the RL agents' objectives \cite{hodossy2023shared}. The exoglove controller is trained until converge, and experiments are conducted to validate the learned exoglove behavior in terms of finger biomechanics and succssess rate, defined in Sec. \ref{sec:manipulation-task}.

\subsection{Validation of exoglove biomechanics in simulation}
The actuation of the exoskeleton glove exerts a force on the index and middle fingers to support the contraction and extension of the finger during grasping. Given the actuation mechanism, the distance between the PIP joint and the wrist is measured as an indicator of glove support.

Fig. \ref{fig:glove-mech} shows the biomechanical assistance provided by the exoskeleton glove. In the healthy case, the distance from the PIP joint to the wrist is first reduced during the closure of the palm, and then increased to hold the grasp gesture. Whereas in the case with simulated muscle weakness, the hand fails to achieve the same contraction due to the lack of muscle strength, causing the hand to drop the object. In contrast, the learned exoglove controller is able to restore an actuation pattern that matches to that of the healthy case, in terms of joint angle changes.

\begin{figure}
    \centering
    \includegraphics[width=0.9\linewidth]{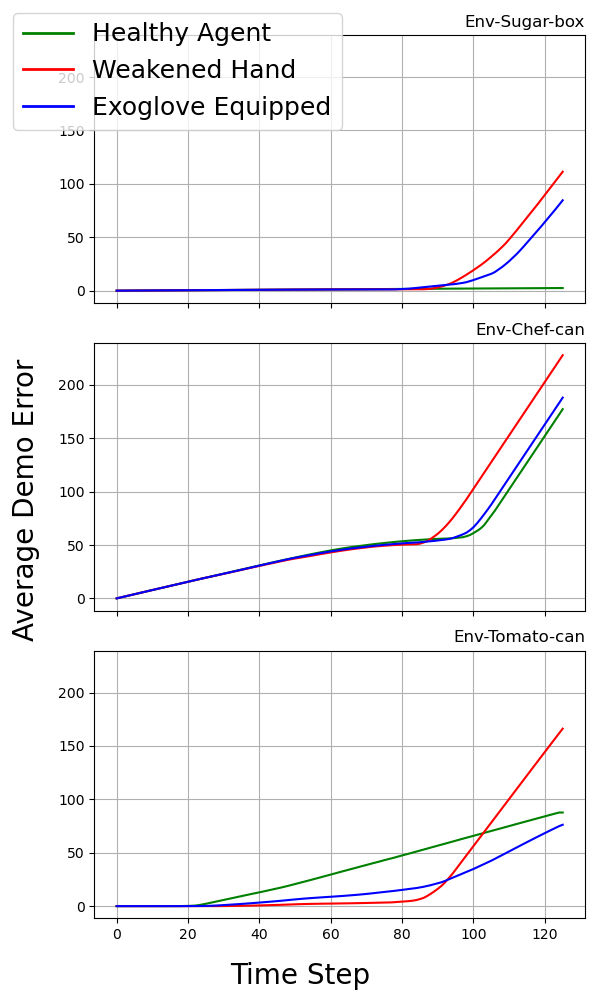}
    \caption{Accumulated error in object manipulation tasks with the exoglove controller’s compensation showing healthy (green) and motor impaired (red) hands, and motor impaired hand with exoglove support (blue). Objects from left to right: chef can, sugar box and tomato can from the YCB objects \cite{calli2015ycb}. The accumulated error illustrates the effectiveness of the exoglove in supporting grasping. The glove controller restores an average of 90.5\% of the original success rate.}
    \label{fig:glove-control}
\end{figure}

Fig. \ref{fig:glove-control} shows the effect of the exoglove controller to support object-centric manipulations, and measured by an error matrix. The error measures the distance between the object's observed position and the reference position. When the deviation is large, defined by an error of 20 cm away from the reference position, a punishment value is added. The rapid rising of accumulated error, thus, indicates that the simulated motor impairments result in a lack of grasping force, rendering the musculoskeletal hand to drop the object after certain time steps, which aligns to the observed joint angle change as in Fig. \ref{fig:glove-mech}. Meanwhile, the learned glove controller was able to capture the hand's behaviour and learned a control strategy to provide suitable assistance for enhanced graspin. These results prove that the learned exoglove controller can capture the subject's joint kinematics and provide suitable support.

\section{Discussion and Conclusion}

In this work, we presented a framework for deriving user-specific reference trajectories using imitation learning, focusing on hand joint kinematics extracted from video data of a healthy subject. These trajectories served as a foundation for reconstructing kinematics and muscle dynamics in silico, where a hand impairment was simulated. We introduced a tendon-driven soft exoglove control strategy designed to compensate for impaired hand functionality. This approach leveraged the healthy reference as a baseline for optimizing the exoglove's control, enabling personalized assistance.

The current approach is not without its limitations. A primary challenge involves translating the simulated outcomes into real-world applications, which remains inherently difficult for most RL-based models. While the state space in simulation can be accurately observed through vision-based extraction algorithms, real-world scenarios introduce complexities such as visual occlusions and partially observable inputs, posing challenges to the exoglove controller's adaptability. Although prior research has explored distilling simulation-trained policies into vision-based policies, the unpredictable nature of real-world environments, particularly in the case of impairments, continues to present a challenge for translating robust and reliable control strategies.

Future work will focus on addressing the deployment of this framework in real-world scenarios, validating the controller with muscle-weakened hands. Additional efforts will also be directed towards bridging the gap between simulation and real-world implementation, increasing generalizability and ensuring the adaptive controller's robustness in dynamic environments across multiple users, objects and tasks.


\bibliographystyle{ieeetr}
\bibliography{reference}

\end{document}